\documentclass[runningheads]{llncs}

\usepackage{accv}

\usepackage{comment}
\usepackage[utf8]{inputenc}  
\usepackage[T1]{fontenc}
\usepackage{CJKutf8}

\usepackage{accvabbrv}

\usepackage{graphicx}
\usepackage{booktabs}
\usepackage{amsmath}
\usepackage{bm}
\usepackage[accsupp]{axessibility}  

\usepackage[pagebackref,breaklinks,colorlinks,citecolor=accvblue]{hyperref}

\begin{document}
\begin{CJK}{UTF8}{gbsn}
\title{Manipulating Vehicle 3D Shapes through Latent Space Editing} 


\author{JiangDong Miao\inst{1} \and
Tatsuya Ikeda\inst{2} \and
Bisser Raytchev\inst{1} \and
Ryota Mizoguchi\inst{1} \and
Takenori Hiraoka\inst{1} \and
Takuji Nakashima\inst{1} \and
Keigo Shimizu\inst{3} \and
Toru Higaki\inst{1} \and
Kazufumi Kaneda\inst{1}
}

\authorrunning{J. Miao et al.}

\institute{Hiroshima University, Graduate School of Advanced Science and Engineering, Hiroshima, Japan 
\and
Hiroshima University, School of Informatics and Data Science , Hiroshima, Japan \\
\and
Mazda Motor Corporation, Hiroshima, Japan\\
}

\maketitle

\begin{abstract}
Although 3D object editing has the potential to significantly influence various industries, recent research in 3D generation and editing has primarily focused on converting text and images into 3D models, often overlooking the need for \emph{fine-grained control} over the editing of existing 3D objects. This paper introduces a framework that employs a pre-trained regressor, enabling continuous, precise, attribute-specific modifications to both the stylistic and geometric attributes of vehicle 3D models. Our method not only preserves the inherent identity of vehicle 3D objects, but also supports multi-attribute editing, allowing for extensive customization without compromising the model’s structural integrity. Experimental results demonstrate the efficacy of our approach in achieving detailed edits on various vehicle 3D models.
\end{abstract}

\section{Introduction}
\label{sec:intro}

The recent advancements in 3D modeling technologies have the potential to significantly influence various industries, especially design-related industries like vehicle design, where style and customization play pivotal roles. 
However, current 3D editing methods which mainly focus on converting/fusing text or images into 3D models, often lack the granularity required for detailed modifications based on existing models. This gap highlights the need for more mature tools that can provide designers with precise control over the editing process, particularly to enhance the stylistic and geometric aspects of the relevant 3D objects' design.

The primary objective of this research is to achieve fine-grained editing for 3D vehicle models, for example to be able to change in a continuous manner the height/width of the vehicle's roof/hood/cabin etc. (\emph{geometric editing}), or similarly to change some stylistic attributes, e.g. make a certain powerful-looking car to be a little bit more delicate, or look more luxurious, etc. (\emph{stylistic editing}). Note that although in this paper we focus on the geometric/stylistic editing of vehicles, the proposed framework is very general and can be applied to any other categories of 3D objects of interest. As industries increasingly rely on 3D visualization and modeling, especially in fields such as vehicle design, quickly previewing a 3D vehicle model after certain geometric/stylistic attributes are changed can not only timely observe whether the changed model conforms to some desirable aesthetic requirements, but also facilitate early aerodynamic research (for example, Arechiga et al. have proposed a model \cite{drag-guided-vehicle} that can generate new vehicle images when simultaneously minimizing their predicted drag coefficients). 

Existing 3D editing approaches have mainly focused on fusing the text or image information into 3D models. Text-guided 3D editing frameworks like DreamEditor, FocalDreamer \cite{dreamedit,focaldreamer} can provide precise modifications while maintaining consistency outside the edited areas. 3DNS \cite{3dNS} proposed the first method for efficient interactive editing of signed distance functions. SKED \cite{SKED} proposed a technique that can edit 3D shapes represented by NeRF based on both text and sketches. Tip-Editor \cite{tipeditor} enables accurate control of the appearance of 3D Guassian splatting represented 3D objects. But text description has its inherent limitations, and sketches also cannot provide fine-grained detail editing guidance.

\begin{figure}[tb]
  \centering
  \includegraphics[height=4.3cm]{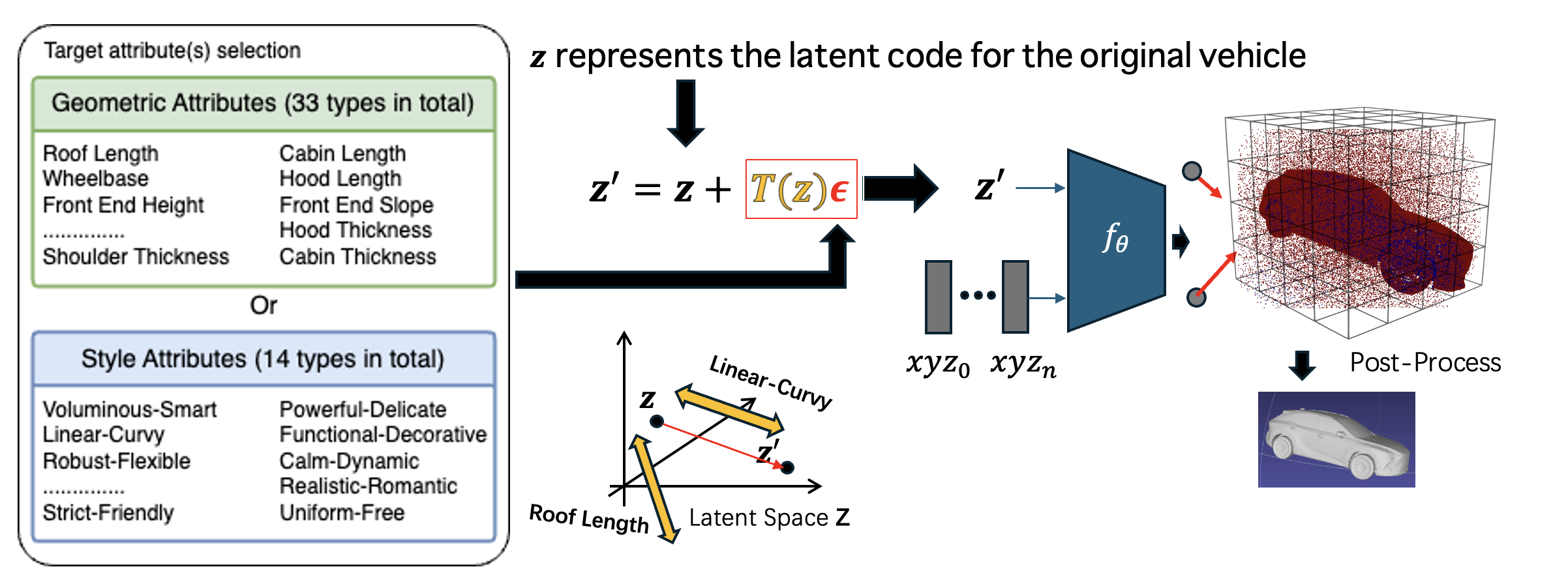} 
  \caption{Overview of the editing flow for the 3D vehicle model proposed in this paper.
  During the editing process, users can specify a parameter (either geometric or stylistic) they wish to edit, along with the desired intensity of the edit. This parameter, representing the editing requirements, is then fed into the trained latent editor. The module adjusts the original latent code of the vehicle model according to the specified editing requirements and returns a new latent code. The updated latent code is input into the pre-trained DeepSDF, which maps it to signed distance function (SDF) values. These SDF values are subsequently processed through the marching cubes algorithm to produce the new 3D vehicle model. 
  }
  \label{fig:overview-edit}
\end{figure}

\textbf{Contributions} In this paper, we introduce a novel framework, an overview of which is depicted in Figure \ref{fig:overview-edit} and the details are given in section 3. The framework is designed for precise editing within the latent space of 3D vehicle models. This framework leverages on the characteristics of the DeepSDF model \cite{DeepSDF}, which is able not only to reconstruct 3D vehicle objects, but also to obtain latent representations suitable for the continuous editing of different attributes of the objects. We achieve this by training a 3D object regressor based on the latent code representations. Inspired by the method proposed by Zhuang \etal's framework \cite{ZhuangEdit} for 2D image editing, our framework introduces a novel approach to the fine-grained editing of 3D vehicle models using the latent space. Unlike many existing 3D generation and editing methods, this framework enables direct manipulation of real 3D objects based on detailed attribute data, allowing for more precise control over the editing process. This approach extends the capabilities of the DeepSDF model beyond simple reconstruction, enabling  geometry and style-related dynamic adjustments tailored to specific editing requirements.
Numerous experimental results demonstrate the efficacy of our approach in achieving detailed fine-grained control over the edits on various vehicle 3D models. Furthermore, our method is not only able to preserve the inherent identity of the edited 3D vehicle objects, but also supports simultaneous multi-attribute editing, allowing for extensive customization without compromising the model’s structural integrity.

\section{Related Work}

\subsection{3D Generation Models}

In the domain of 3D model generation, various data structures have been employed to represent 3D shapes, encompassing point clouds, voxels, meshes, and neural fields. Notably, Neural Radiance Fields (NeRF) \cite{Nerf} and their improved versions \cite{nerfw,mid-nerf} have attracted significant attention for their ability to generate high-quality, multi-view images of 3D models. These methods have set a new standard for photorealism in view synthesis.

A different approach, 3D Gaussian Splatting \cite{3Dgaussians}, uses distributed Gaussian kernels for volumetric rendering. This technique is particularly advantageous in dynamic environments and real-time applications due to its efficiency in scene synthesis.

Parallel to these developments, diffusion models have recently emerged as a powerful generative framework in the 3D domain. Methods such as 3D-LDM \cite{3dldm} and Diffusion-SDF \cite{diffusion-sdf} have introduced ways to leverage these models for 3D shape synthesis. However, despite their potential, these approaches generally require extensive datasets for effective training, presenting a significant challenge, especially in specialized domains where data collection is non-trivial (like in the case of vehicle 3D editing on which we focus in this paper, where only a very small number of training data might be available). Furthermore, the training process for these models tends to be slow, limiting their practicality for rapid deployment.

In contrast, DeepSDF \cite{DeepSDF} uses a signed distance function (SDF) to offer a continuous, implicit field representation, which stands apart from the image-centric outputs of NeRFs and 3D Gaussian Splatting. Unlike these methods that primarily synthesize images, DeepSDF directly generates 3D objects through the Marching Cubes algorithm \cite{matchcube}. It leverages SDFs within a high-dimensional latent space, optimizing a unique latent code for each 3D object. This capability allows for precise manipulation and interpolation of shapes, translating subtle variations in latent space into meaningful geometric modifications in the 3D model. Thus, DeepSDF enables more effective complex shape modifications compared to other contemporary methods and this is the reason we have adopted a modified version of it in our framework to generate the latent space representations of the 3D vehicle models on which the fine-grained attribute editing is performed.

\subsection{Semantic Editing}
Semantic editing has been extensively explored within the realm of 2D imagery, yet its extension to 3D modeling—particularly for detailed attribute manipulation in specialized fields such as automotive design—remains in a nascent stage. Semantic editing techniques in the 2D domain have demonstrated their efficacy in facilitating precise modifications of image attributes under specific conditions, predominantly employing generative adversarial networks (GANs) \cite{gan,stylegan2,PGGAN} and variational autoencoders (VAEs) \cite{VAE}. These techniques enable transformative edits such as altering the seasonal depiction from summer to winter or modifying hair color from black to white \cite{face-edit}. However, although approaches like those introduced in \cite{GANcontrol,Gan-unsup} can perform unsupervised editing in latent spaces, they tend to alter the image’s identity during the process.

As mentioned in \cite{survy3}, the historical context of \emph{facial attribute editing} extends back to the 1990s \cite{past1,past2}. And actually, editing facial attributes shares substantial parallels with editing 3D vehicle models. Both domains exhibit varying levels of abstraction and can be categorized into local and global attributes. For instance, specific facial attributes such as hair length or eye color target isolated regions of the face. Conversely, in vehicle models, localized attributes like the hood length predominantly affect the vehicle’s frontal aesthetics without impacting the overall design, similar to how the inclination angles of the front and rear windows relate specifically to their respective positions. Global attributes, such as age in faces or wheelbase and total height in vehicles, influence a broader range of components.

As elucidated in \cite{survyfacialedit1,survyfacialedit2}, facial editing can be classified into single-attribute and multi-attribute methods. Notably, \cite{age} focuses exclusively on age modification, whereas \cite{glasses} specializes in removing glasses from images. Multi-attribute techniques like relGAN \cite{relGan} can simultaneously alter several features, including hair color.

Among these image editing methods using GAN models, a large part of them utilize the semantic information contained in the latent space of the GAN model. For example, Tewari et al. \cite{styleRig} leveraged StyleGAN’s latent space to transpose external attributes from source to target images, and \cite{InterFaceGAN} employed the latent spaces of various GAN models for facial attribute editing. Additionally, TunaGAN \cite{TunaGANIG} integrates an auxiliary network to control StyleGAN’s latent space. Zhuang et al. \cite{ZhuangEdit} proposed a framework for editing 2D images within the latent space of a GAN model, using a pre-trained regressor to achieve fine-grained control over various attributes. Specifically, adjusting the latent code in a designated direction within a suitable space can induce significant alterations in the resultant images. This process involves training models to discern the correct direction and magnitude of latent code adjustments to effectuate the desired attribute changes in the generated objects. The precision and scalability demonstrated by their model in editing facial attributes have attracted considerable interest.

Despite the progress in 2D semantic editing, where techniques like StyleGAN leverage the rich semantic information within latent spaces, adapting these techniques for 3D modeling presents unique challenges and opportunities. In 3D domains, especially those lacking substantial training data, it is uncertain whether the activated latent codes during the training of generative models possess enough semantic information for effective editing. Furthermore, the quality of 3D reconstructions generated by models trained in data-scarce environments remains questionable. Given the complexity of 3D structures and their attributes, the necessity for models that can aptly represent, comprehend, and manipulate detailed information in three-dimensional space is paramount. Although 3D semantic editing, particularly in specialized areas such as vehicle modeling, is still evolving, foundational principles established in 2D image editing could potentially provide a robust framework for future advancements in 3D modeling. This paper explores this promising direction, aiming to bridge the gap between 2D success and 3D potential.

\section{Method}
\subsection{Task Definition}
In this paper, we introduce a model that enables fine-grained control of 3D vehicle models by editing specific attributes defined by the dataset through a pre-trained DeepSDF model. Initially, we train the DeepSDF model using our dataset, which establishes a mapping module \( f(\theta) \) that translates the latent vector codes \( \mathbf{z} \), associated with the xyz coordinates of the 3D points sampled from a 3D object, into SDF values (see Figure \ref{fig:overview-edit}). These latent vectors correspond to each 3D model used during training. We then develop a regressor \( R \) to aid the training of latent code editors. For a pre-defined set of geometry or style-related attributes \( \mathcal{A} = \{A_1, A_2, A_3, \ldots, A_n\} \) — which depend on the dataset and specific interests — we seek to determine precise directions in the latent space of DeepSDF, denoted as \( \mathcal{T} = \{\mathbf{d}_1, \mathbf{d}_2, \mathbf{d}_3, \ldots, \mathbf{d}_n\} \), to modify the corresponding attributes of the original 3D vehicle models as desired.

\subsection{Dataset Collection and Processing}
We collected 180 3D vehicle models along with their corresponding detailed \emph{geometric} specifications, for which the specific parameters are listed in Table \ref{tab:geometric specification}. In addition, we conducted a questionnaire survey to assess the \emph{stylistic} attributes of each vehicle. The survey gathered responses from 180 participants, who rated 65 vehicles in our dataset based on 14 pairs of predefined stylistic attributes (shown in Table \ref{tab:stylistic attributes}). We aggregate these survey results using a weighted scoring method, and the derived scores (normalized to the range [0, 1]) serve as the numerical labels for the final stylistic attributes of each vehicle, to be used in the subsequent training. 

\begin{table}[tb]
  \caption{ List of 33 geometric specification of a vehicle
  }
  \label{tab:geometric specification}
  \centering
  \begin{tabular}{@{}p{3cm}p{3cm}p{5cm}@{}}
    \midrule
    Roof Length & Cabin Length & Front Wheel-L Length\\
    L2 & Wheelbase & Key Line Base Length\\
    Hood Length& Front Overhang & Side Sill Lower Edge Height\\
    RR-OH/FR-OH & H1 & Front Bumper Lower Edge Height\\
    Front End Height & Front End Slope & Key Line Base Height\\
    Belt Line Height & Total Height & Rear Bumper Lower Edge Height\\
    Rear Overhang & Key Line Angle & Front Window Inclination Angle\\
    Cabin Width & Overall Width & Rear Window Inclination Angle\\
    Roof Width & Tread Width & Bumper Lower Edge Width\\
    Cabin Thickness & Roof Thickness & Bumper Upper Edge Width\\
    Shoulder Thickness & Hood Thickness & Bumper Thickness\\
  \bottomrule
  \end{tabular}
\end{table}

\begin{table}[tb]
  \caption{ List of 14 pairs of predefined stylistic attributes
  }
  \label{tab:stylistic attributes}
  \centering
  \begin{tabular}{@{}lll@{}}
    \midrule
    Voluminous-Smart & Powerful-Delicate & Linear-Curvy\\
    Functional-Decorative & Robust-Flexible & Calm-Dynamic\\
    Realistic-Romantic & Elegant-Cute & Sophisticated-Youthful\\
    Luxurious-Approachable & Formal-Everyday & Strict-Friendly\\
    Uniform-Free & Special-Everyday\\
  \bottomrule
  \end{tabular}
\end{table}

\subsection{Proposed Approach}

\subsubsection{Overview}

\begin{figure}[t]
  \centering
  \includegraphics[height=3.5cm]{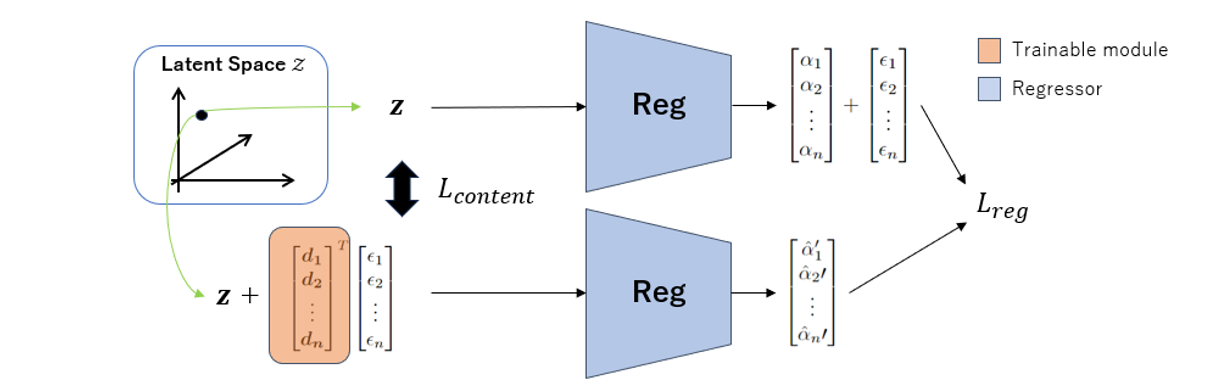}
  \caption{The Training Strategy for the Latent Code Editor.
  In our training pipeline, the latent code \( \mathbf{z} \) is sampled from a normal distribution. The elements \( \{\mathbf{d}_1, \mathbf{d}_2, \ldots, \mathbf{d}_n\} \) represent trainable modules that denote directions in the latent space, and \( \epsilon \) denotes the attributes to be edited along with their transformation magnitudes. The attribute values of the original and transformed latent codes are predicted by a pre-trained regressor. The regression loss \( L_{\text{reg}} \) and the identity loss \( L_{\text{context}} \) are employed to update the parameters. The details of the training flow are explained in \cref{sec:trainflow}.
  }
  \label{fig:training}
\end{figure}

In 
\cref{fig:training}
we present the training framework for 3D vehicle model editing developed in this paper, which consists of a latent code editor and a pre-trained regressor. The primary objective of the training is to identify modifications within the latent space that enable the transformation of the new 3D vehicle model, represented by \( \mathbf{z}' \), to achieve desired changes in the target attributes relative to the original model.

Following the experience from the 2D image editing domain \cite{ZhuangEdit}, the Editor, initialized with random parameters, modifies the input latent code \( \mathbf{z} \) based on the specified transformation magnitude \( \epsilon \). The pre-trained regressor predicts attribute values for both \( \mathbf{z} \) and \( \mathbf{z}' \) before and after the editing process. In our experiments, all attribute values for style and geometry are normalized to the range [0, 1]. The attribute values for the original 3D model, represented by \( \mathbf{z} \), are denoted as \( \alpha = R(\mathbf{z}) \). For the modified latent code \( \mathbf{z}' \), the attribute values of the 3D model are denoted as \( \hat{\alpha}' = R(\mathbf{z}') \). Additionally, the pseudo ground truth, represented as \( \alpha' \), is defined by \( \alpha' = \alpha + \epsilon \). This pseudo ground truth \( \alpha' \) serves as a target in training to guide the modifications toward the desired attribute changes.

Specifically, the original input \( \mathbf{z} \) during training is a latent vector sampled from a normal distribution, while \( \epsilon \) is randomly sampled from a uniform distribution over the interval \([-1, 1]\), with the constraint that \(0 \leq (\alpha + \epsilon) \leq 1\).

\subsubsection{Position Enhanced DeepSDF}

\begin{figure}[t]
  \centering
  \includegraphics[height=6cm]{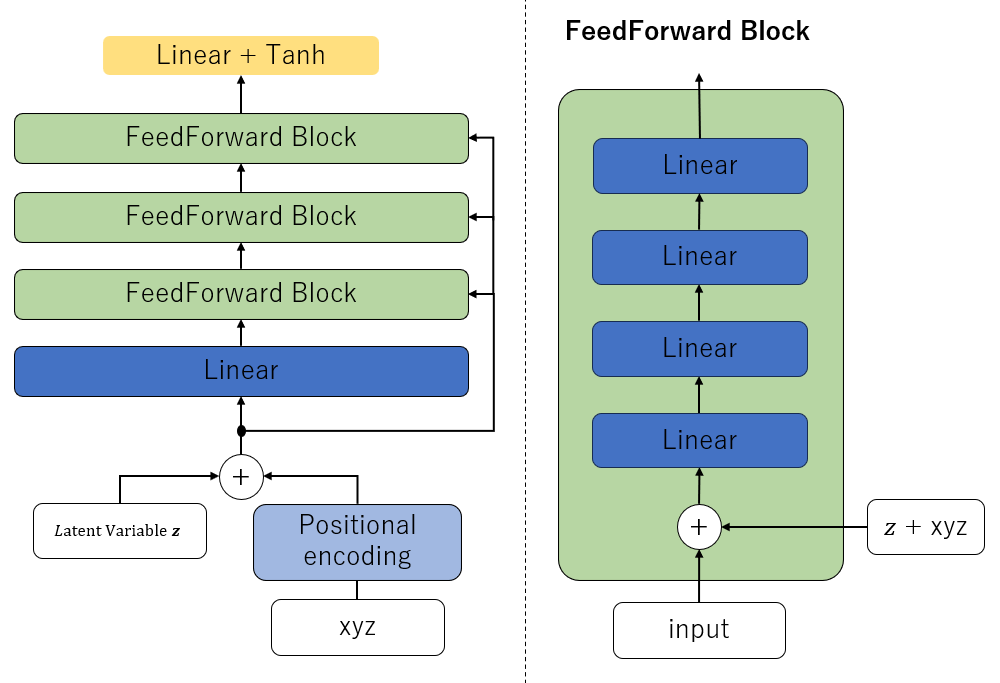}
  \caption{Position Enhanced DeepSDF
  }
  \label{fig:deepsdf}
\end{figure}

\cref{fig:deepsdf} illustrates the training workflow of our modified DeepSDF model, enhanced to utilize position embedding. Unlike the original DeepSDF model, which requires the $xyz$ coordinates of the 3D sampling points as input during training, our approach, inspired by the NeRF model, incorporates position embedding before feeding the $xyz$ coordinates into the model. This modification significantly enhances the model’s capability to recognize features to coordinate positions. Our experimental results demonstrate that integrating NeRF’s position embedding technique notably improves the generative performance of the model. The position embedding function is similar to that which NeRFs use and can be expressed as in Eq. (\ref{eq:pos-emb}) below:

\begin{equation}
  \gamma(p) = (\sin(2^0\pi p), \cos(2^0\pi p), ... , \sin(2^L\pi p), \cos(2^L\pi p)).
  \label{eq:pos-emb}
\end{equation}

Note that while the NeRF model \cite{Nerf} facilitates the generation of images from various perspectives of 3D objects, the DeepSDF model can directly construct the 3D object itself using the marching cubes algorithm. During training, DeepSDF optimizes a specific latent code for each 3D model, which is crucial for our subsequent latent-space-based editing steps. The model architecture, consisting of a stack of linear layers, offers simplicity and ease of training. These features are the main reasons why we prefer using the modified version of DeepSDF described above, instead of other 3D generation models (which also would require a much larger dataset for training than the one we have).

\subsubsection{Latent Regressor}
In the proposed training framework, we employ a pre-trained regressor to make fine-grained predictions and edits on the attributes of 3D objects. Given the absence of a robust feature extractor for 3D objects, we utilize the latent codes optimized for each 3D vehicle model from the previous training phase with the modified DeepSDF model. To predict style and geometric attributes, we trained a simple 4-layer Multilayer Perceptron (MLP) using these latent codes as input and the normalized scores for the geometric/style attributes for our dataset as output. The ability to leverage these optimized latent codes is another reason that we use DeepSDF for this framework.

\subsubsection{Latent code Editor}
\paragraph{\textbf{MLP-based Editor}}
In this part, we designed a latent code editor based on MLP, which can output the edited latent code according to the original latent code and the strength of attribute editing:

\begin{align}
    \mathbf{v} &= \text{MLP}_1(\mathbf{z}), \\
    \mathbf{u} &= \frac{\mathbf{v}}{\|\mathbf{v}\|} \lambda, \\
    \mathbf{z}' &= \mathbf{z} + \text{MLP}_2(\epsilon \cdot \mathbf{u}),
\end{align}
where \(\text{MLP}_1\) and \(\text{MLP}_2\) are specific implementations of the Multi-Layer Perceptron (MLP) layers, \(\lambda\) is a scaling factor hyper-parameter, and \(\|\mathbf{v}\|\) denotes the L2 norm of \(\mathbf{v}\).

\paragraph{\textbf{KAN-based Editor}}
In recent times, the KAN model has attracted widespread attention. Therefore, we also designed a latent code editor based on the KAN model. Kolmogorov-Arnold Networks (KANs) \cite{KAN}, inspired by the Kolmogorov-Arnold Representation Theorem, have recently gained significant attention. This theorem suggests that any multivariate continuous function can be decomposed into univariate functions and a summing operation, a principle directly integrated into the KAN structure.

Unlike traditional Multi-Layer Perceptrons (MLPs) that use fixed activation functions at the nodes, KANs employ learnable functions on the edges of the graph, enhancing the model's adaptability and making them a promising alternative to MLPs. The computational formula for a KAN, involving \(k\) learnable activation functions, is shown below:

\begin{equation}
    KAN(\mathbf{x}) = \left(\Phi_k \circ \Phi_{k-1} \circ \cdots \circ \Phi_1\right)(\mathbf{x})
    \label{eq:kan}
\end{equation}

In our framework, we also utilize a latent code editor based on KAN. The editor processes the latent codes \(\mathbf{z}\) and attribute perturbations \(\epsilon\), resulting in updated latent codes \(\mathbf{z}'\) as follows:
\begin{align}
    \mathbf{v} &= \text{KAN}_1(\mathbf{z}), \\
    \mathbf{u} &= \frac{\mathbf{v}}{\|\mathbf{v}\|} \lambda, \\
    \mathbf{z}' &= \mathbf{z} + \text{KAN}_2(\epsilon \cdot \mathbf{u}),
\end{align}
where \(\text{KAN}_1\) and \(\text{KAN}_2\) are specific implementations of the Kolmogorov-Arnold Network layers, \(\lambda\) is a scaling factor hyper-parameter, and \(\|\mathbf{v}\|\) denotes the L2 norm of \(\mathbf{v}\).

\subsubsection{Training Flow}
\label{sec:trainflow}
Having outlined the necessary prerequisites, we now detail the overall training procedure, which is straightforward. Initially, \( \mathbf{z} \) is sampled randomly from a normal distribution, and the modification of the target attribute \( \epsilon \) is drawn from a uniform distribution. Upon inputting \( \mathbf{z} \) and \( \epsilon \) into the latent editor, 
the edited latent code \( \mathbf{z}' \) is obtained. Subsequently, \( \mathbf{z} \) and \( \mathbf{z}' \) are fed into a pre-trained regressor to predict their respective attribute values \( \alpha \) and \( \alpha' \). The computation of the training loss has two parts, as shown below: 


\begin{equation}
    \underset{\mathcal{T}}{\rm{min}}\mathcal{L} \triangleq \lambda_1 \mathcal{L}_{reg} + \lambda_2 \mathcal{L}_{content}. 
    \label{eq: t-objective}
\end{equation}
The objective function, represented in Equation (\ref{eq: t-objective}), aims to minimize the total loss \(\mathcal{L}\), which is a weighted sum of the regressor loss \(\mathcal{L}_{\text{reg}}\) and the content loss \(\mathcal{L}_{\text{content}}\). The weights, \(\lambda_1\) and \(\lambda_2\), are hyperparameters that balance the influence of each loss component on the overall training process. In all experiments reported in the paper, we set \(\lambda_1 = 1\) and \(\lambda_2 = 8\) which have been found empirically to work best.

In the objective function above, the first part, \( \mathcal{L}_{reg} \), assesses whether the attributes of the 3D object represented by \( \bm{z}' \) have changed as desired, using the binary cross entropy. The respective formula is shown in Eq. (\ref{eq: reg_loss}) below.

\begin{equation}
    \mathcal{L}_{reg} = E_{\bm{z}\sim\mathcal{Z}, \bm{\epsilon}\sim\mathcal{D}_{\epsilon}}[-\hat{\bm{\alpha}}^{\prime}\log \bm{\alpha}^{\prime} - (1-\hat{\bm{\alpha}}^{\prime})\log (1-\bm{\alpha}^{\prime})]
    \label{eq: reg_loss}
\end{equation}

The second part, \( \mathcal{L}_{content} \), ensures that the latent code \( \bm{z} \) alterations do not affect the identity style of the corresponding 3D vehicle model. The loss formula for this part is provided in Eq. (\ref{eq: content_loss}).

\begin{equation}
    \mathcal{L}_{content} = E_{\bm{z}\sim\mathcal{Z},\bm{z}^\prime\sim\mathcal{Z}^\prime|\bm{z}} \underset{i\in\mathcal{D}_{\rm{content}}}{\sum} ||\bm{z}^\prime - \bm{z}||_2^2\label{eq: content_loss}
\end{equation}

\section{Experimental Results}
\subsection{3D Reconstruction Quality}

\begin{figure}[tb]
  \centering
  \begin{subfigure}{0.45\linewidth}
    \includegraphics[height=3cm]{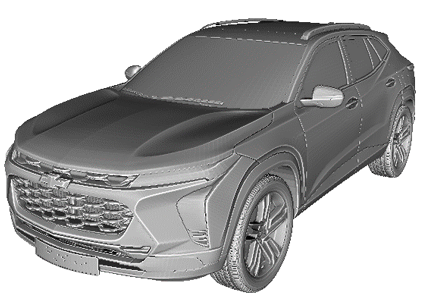}
    \caption{The ground truth of 3D vehicle model}
    \label{fig:vehicle-gt}
  \end{subfigure}
  \hfill
  \begin{subfigure}{0.45\linewidth}
    \includegraphics[height=3cm]{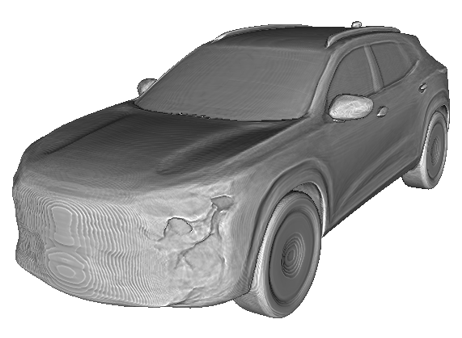}
    \caption{Reconstruction result}
    \label{fig:vehicle-re}
  \end{subfigure}
  \caption{3D Reconstruction quality}
  \label{fig:quality-re}
\end{figure}

In our refinement of the DeepSDF framework, we implemented a position enhanced version of DeepSDF aimed at improving the detail and accuracy of 3D vehicle model reconstructions. The ground truth of a 3D vehicle model and our reconstruction result is shown in  \cref{fig:quality-re}. 
Based on qualitative assessments of the reconstructed models, we assert that our enhanced DeepSDF framework capably reconstructs the 3D vehicle models within our dataset. The fidelity and accuracy of these reconstructions meet the prerequisites required for subsequent fine-grained editing tasks. This level of performance establishes a solid foundation for the next phase of our research, which involves detailed and precise modifications of the 3D vehicle geometries.

\subsection{Position Embedding Improved DeepSDF}
\begin{figure}[tb]
  \centering
  \begin{subfigure}{0.45\linewidth}
    \includegraphics[height=3cm]{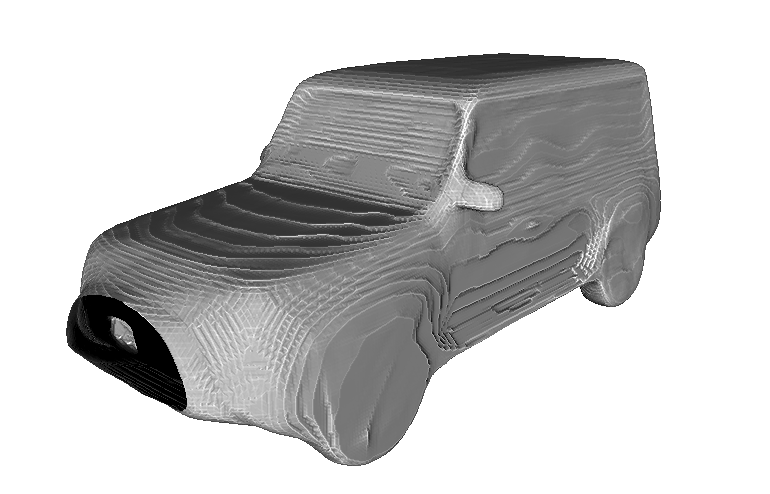}
    \caption{Without position embedding}
    \label{fig:without-pos}
  \end{subfigure}
  \hfill
  \begin{subfigure}{0.45\linewidth}
    \includegraphics[height=3cm]{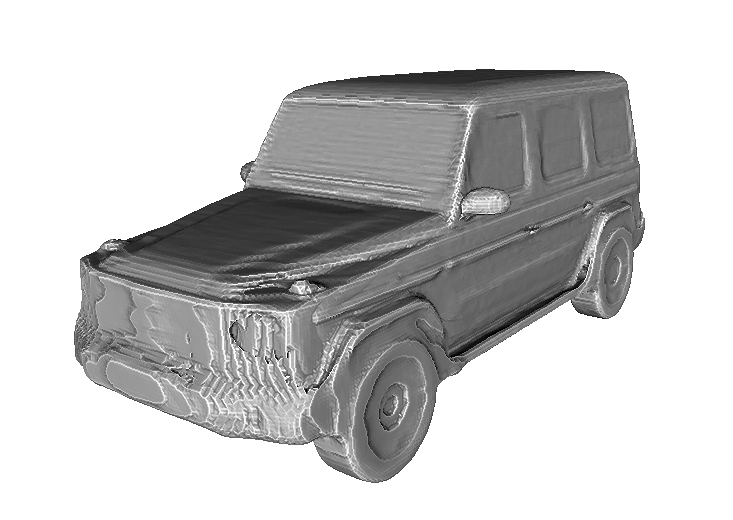}
    \caption{With position embedding}
    \label{fig:with-pos}
  \end{subfigure}
  \caption{The improvement of position embedding}
  \label{fig:compare-pos}
\end{figure}

In our enhancement of the DeepSDF implementation, which draws inspiration from the NeRF model discussed previously, we incorporated position embedding for the $xyz$ coordinates that represent spatial locations. As demonstrated in \cref{fig:compare-pos}, we compare the 3D reconstruction performance of DeepSDF with and without the integration of position embedding. Maintaining consistency in other aspects of the model’s architecture and training both variants for 4000 epochs, the results indicate that the inclusion of position embedding significantly improves the model’s ability to capture finer details of the 3D objects

\subsection{Latent Space Distribution}
Unlike the 2D realm where image generation via GANs often leverages the semantic information embedded within the latent codes, our approach utilizes DeepSDF to reconstruct 3D vehicle models and subsequently extract their corresponding latent codes. One critical question we need to address is whether these latent codes, optimized for 3D reconstruction, are also capable of representing in a smooth and consistent manner the changes in the appearance features of the 3D models as we move with small steps inside the learned embedding space.

\begin{figure}[tb]
  \centering
  \includegraphics[height=6cm]{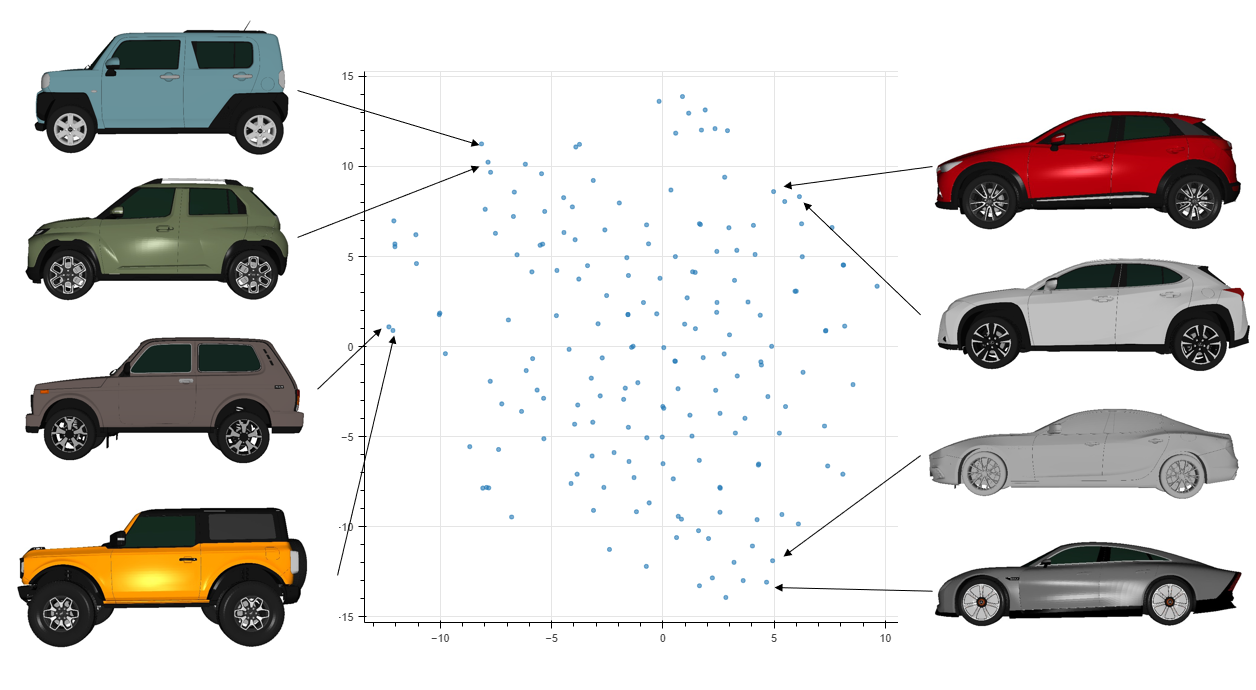}
  \caption{Visualization results based on t-SNE
  }
  \label{fig:tSNE}
\end{figure}

To investigate this, we employed the t-SNE technique \cite{tSNE} to perform dimensionality reduction on the latent codes derived from all training models(\cf \cref{fig:tSNE}). Our observations indicate that the 3D vehicle models corresponding to proximate points in the t-SNE plot exhibit similar appearances, suggesting that the optimized latent codes indeed retain significant semantic information about the objects’ appearances. This finding substantiates the feasibility of applying latent-space editing for 3D models, as it confirms that the latent codes contain meaningful information relevant to the models’ visual traits. This sets a solid experimental foundation for the subsequent latent-space editing processes we intend to explore.

\subsection{3D Vehicle Editing Results}


\begin{figure}[tb]
  \centering
  \begin{subfigure}{0.45\linewidth}
    \centering
    \includegraphics[height=3cm]{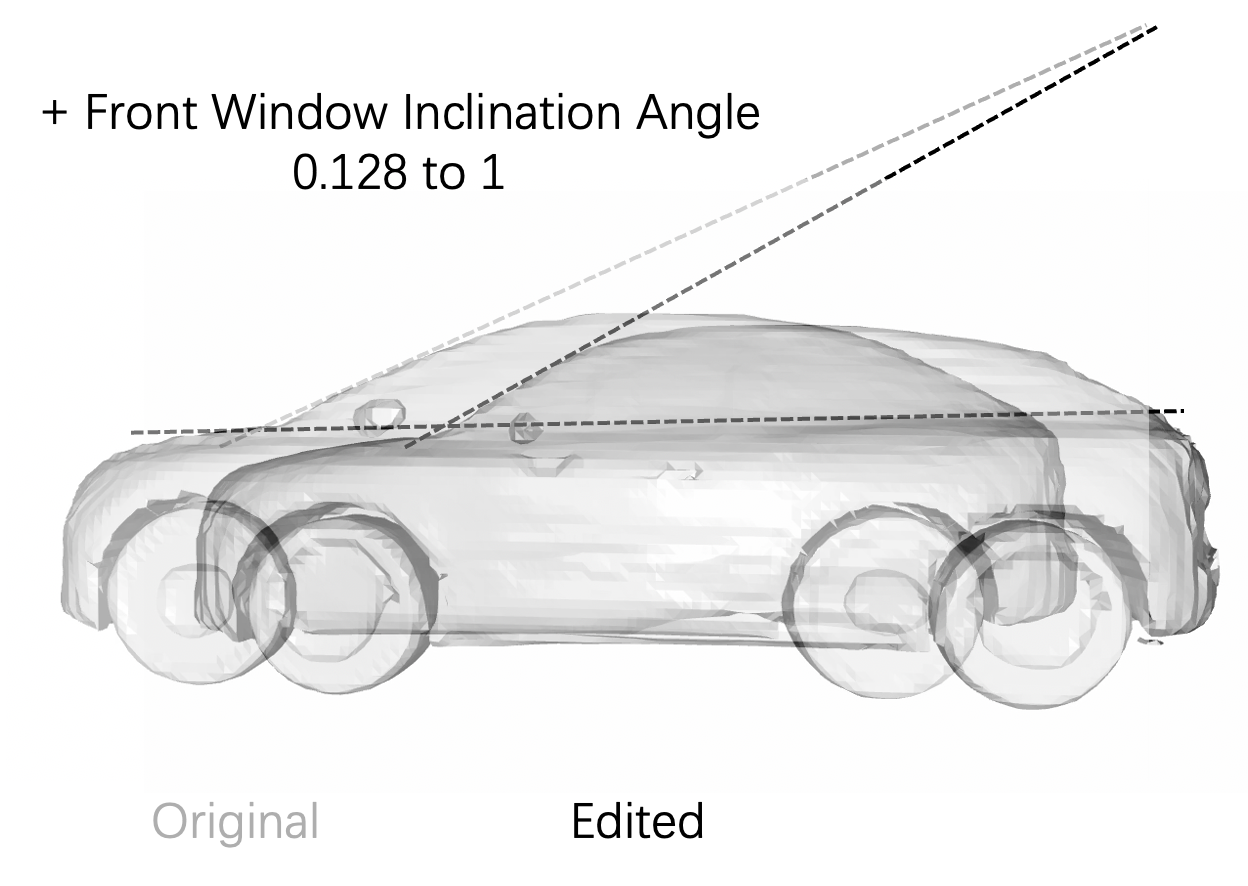}
    \caption{ + Front windows angle}
    \label{fig:geo-FR-Angle}
  \end{subfigure}
  \hfill
  \begin{subfigure}{0.45\linewidth}
    \centering
    \includegraphics[height=3cm]{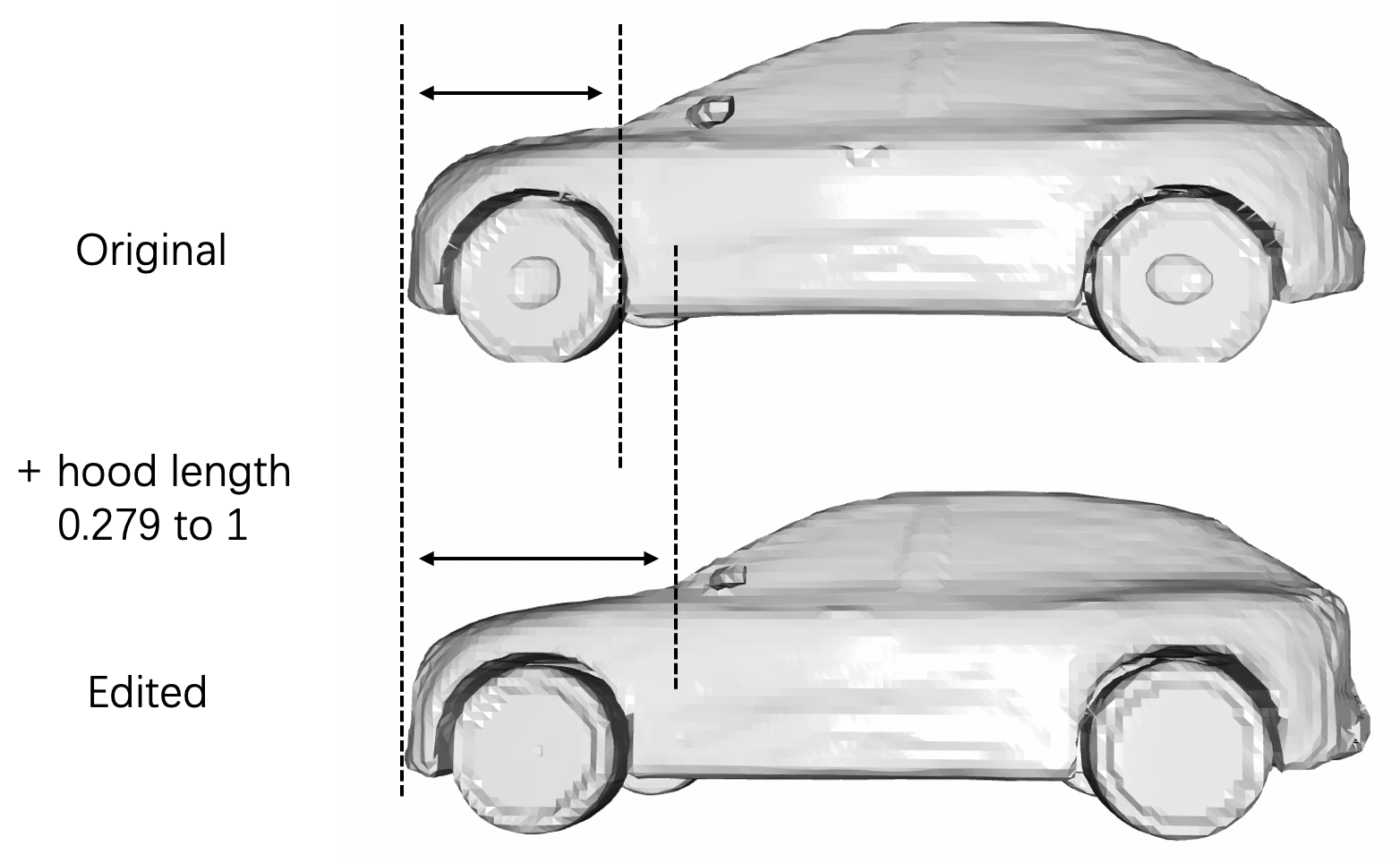}
    \caption{ + Hood length}
    \label{fig:geo-hoodlen}
  \end{subfigure}

  \vspace{0.3cm} 
  
  \begin{subfigure}{0.45\linewidth}
    \centering
    \includegraphics[height=3cm]{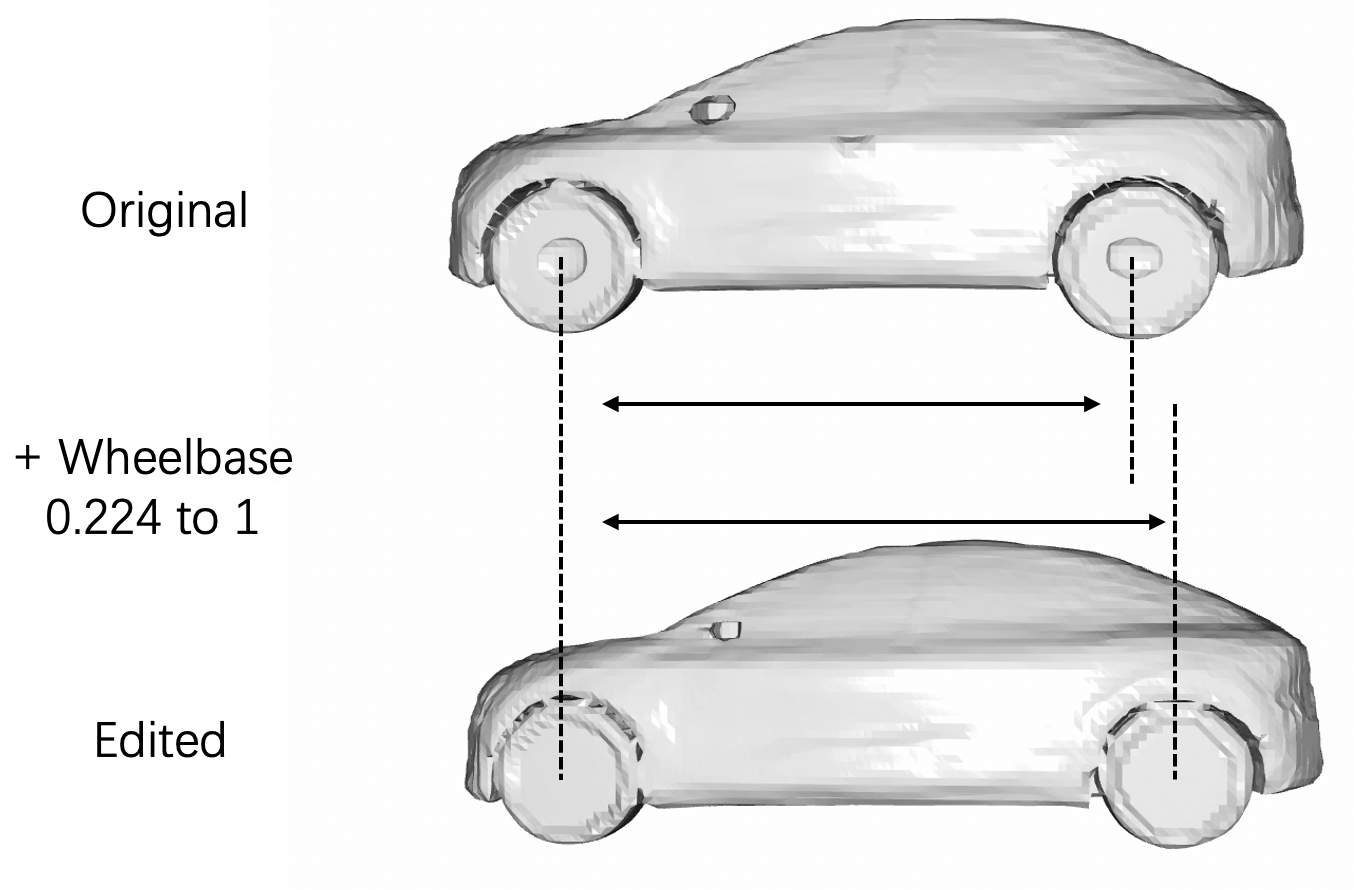}
    \caption{ + Wheelbase}
    \label{fig:geo-wheelbase}
  \end{subfigure}
  \hfill
  \begin{subfigure}{0.45\linewidth}
    \centering
    \includegraphics[height=3cm]{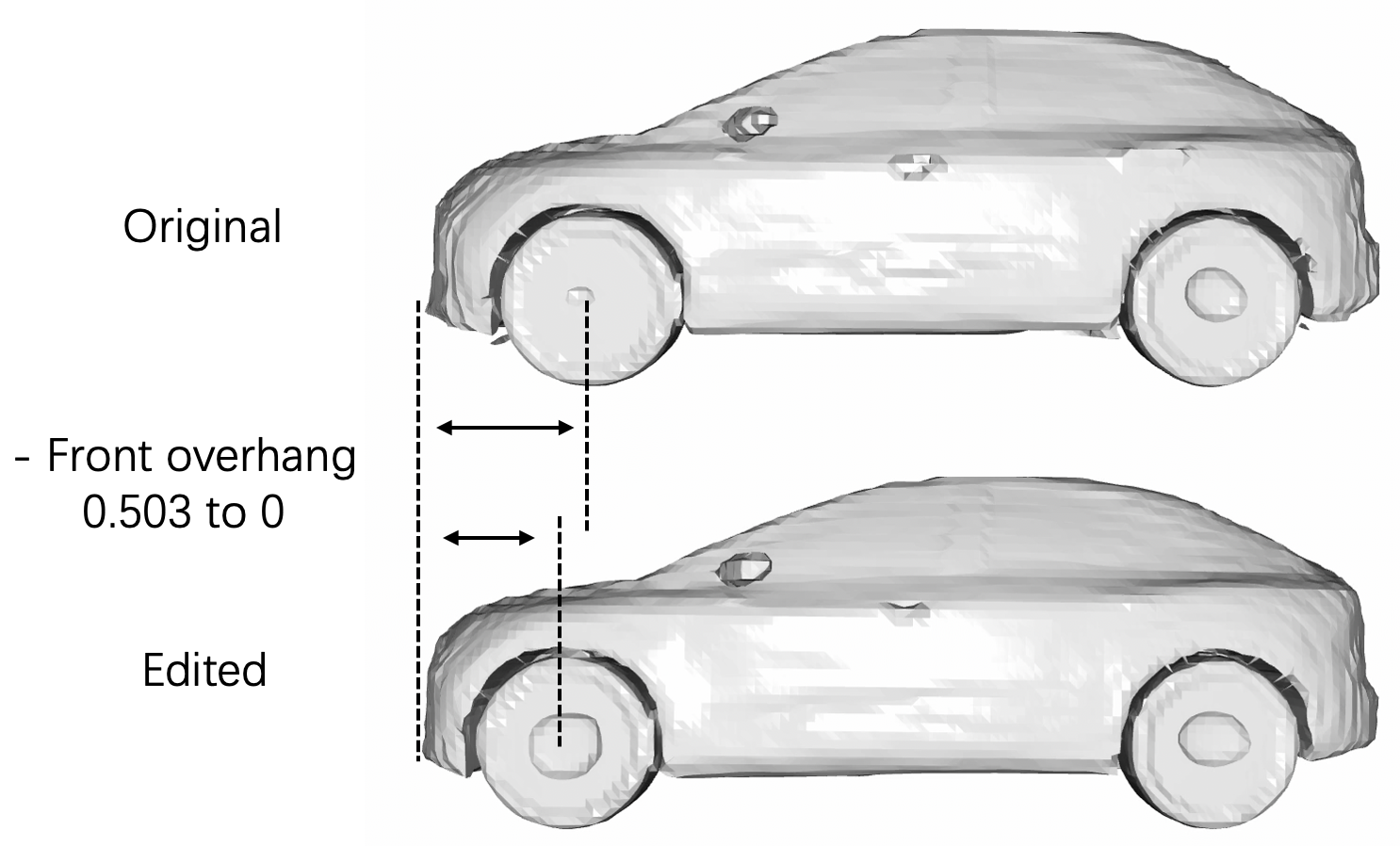}
    \caption{ - Front overhang}
    \label{fig:geo-FR-overhang}
  \end{subfigure}

  \vspace{0.3cm} 

  \begin{subfigure}{0.35\linewidth}
    \centering
    \includegraphics[height=4.5cm]{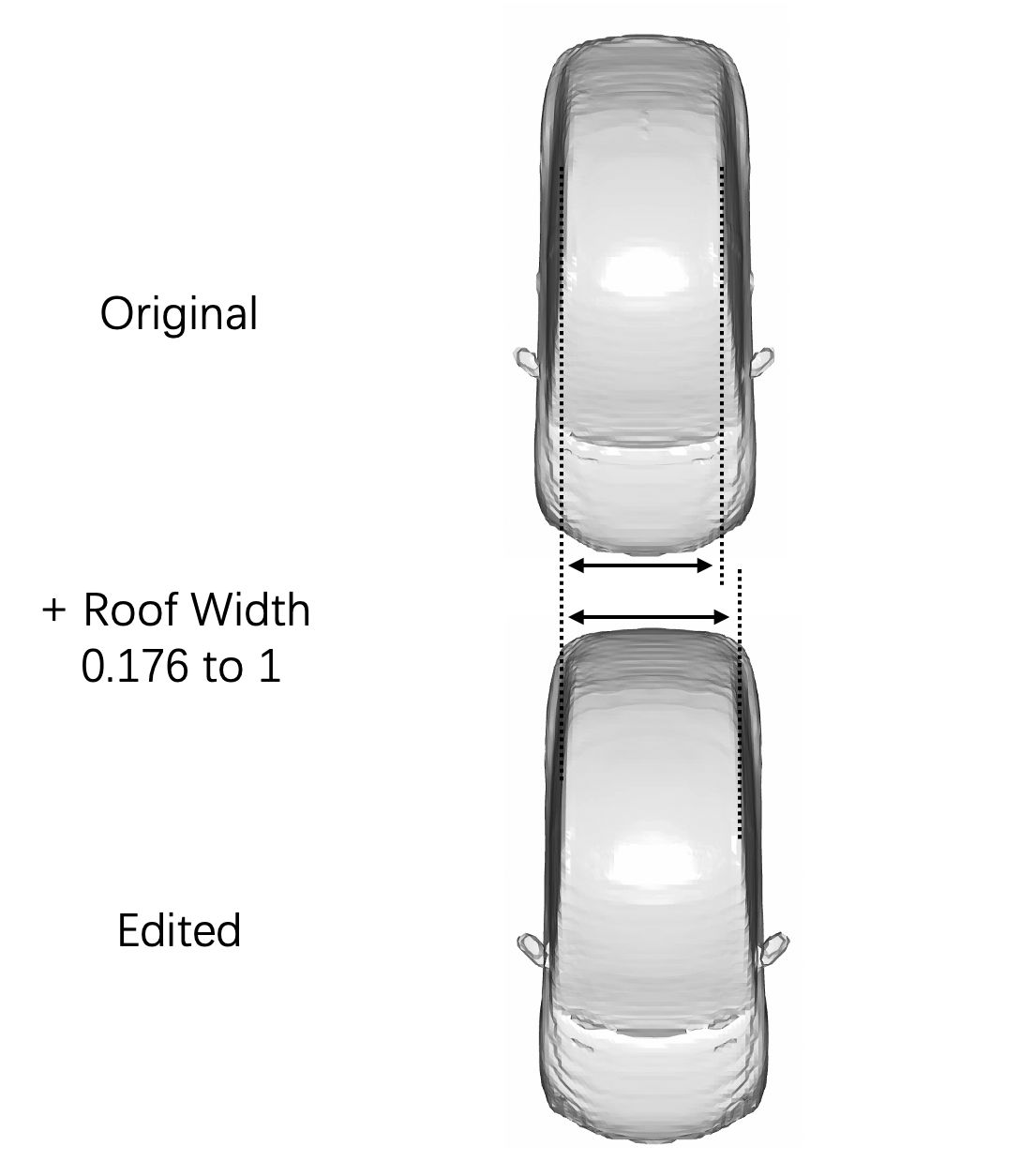}
    \caption{ + Roof width}
    \label{fig:geo-roofwidth}
  \end{subfigure}
  \hfill
  \begin{subfigure}{0.45\linewidth}
    \centering
    \includegraphics[height=3.5cm]{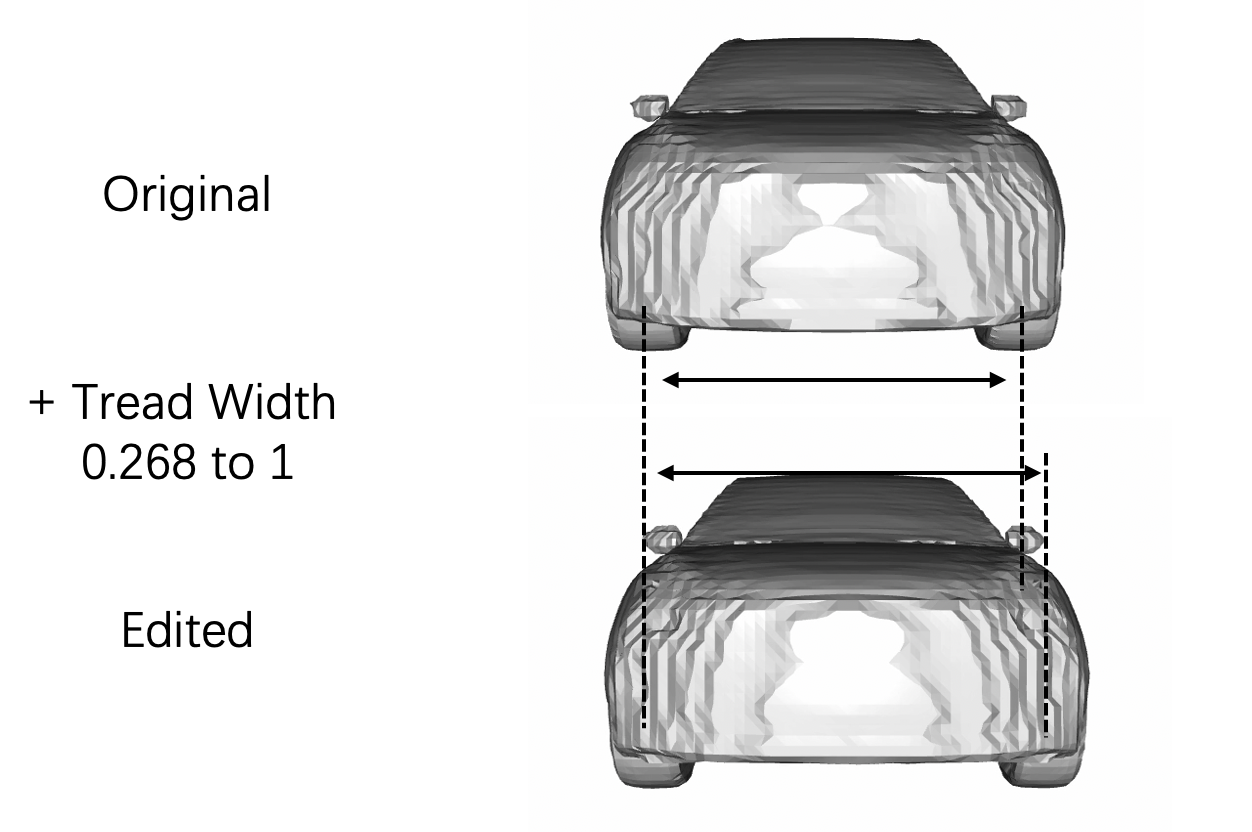}
    \caption{ + Tread width}
    \label{fig:geo-treadwidth}
  \end{subfigure}

  \caption{Single Attribute Editing for Geometric Attributes (1) }
  \label{fig:geo-single}
\end{figure}

\begin{figure}[tb]
  \centering
  
  \begin{subfigure}{0.55\linewidth}
    \centering
    \includegraphics[height=2cm]{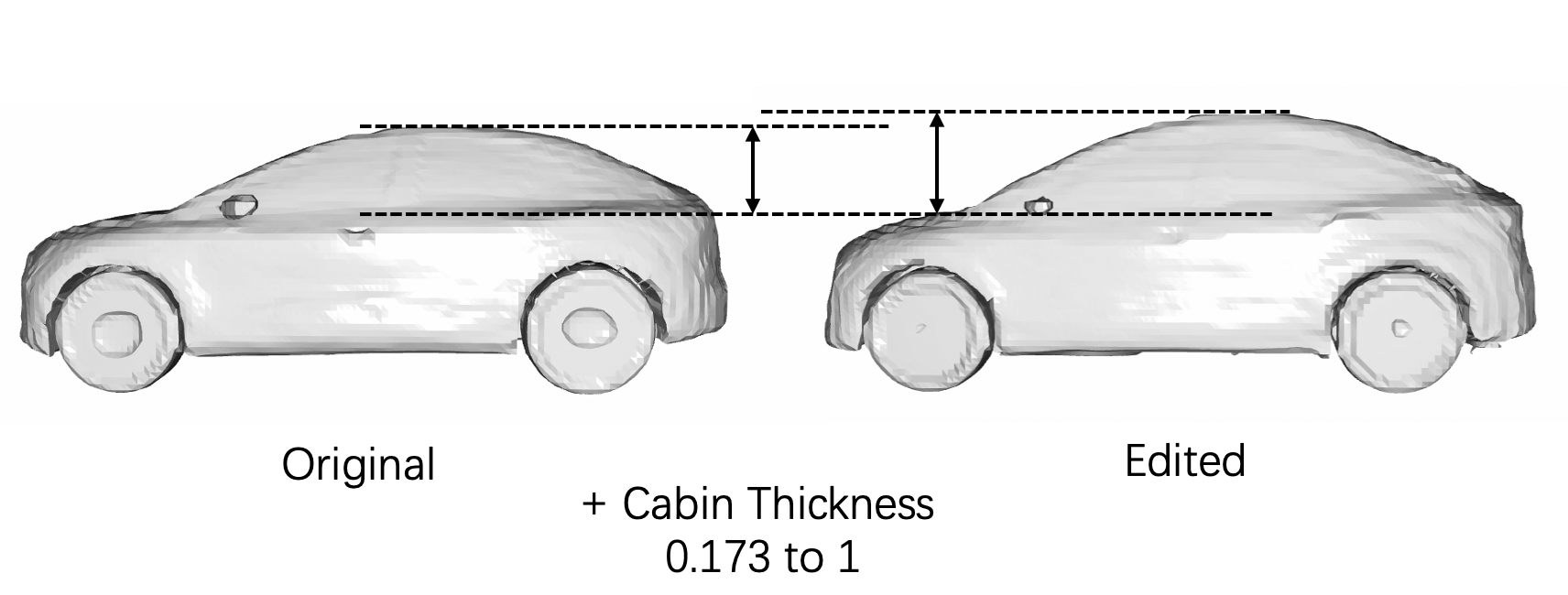}
    \caption{ + Cabin thickness}
    \label{fig:geo-cabinthickness}
  \end{subfigure}
  \hfill
  \begin{subfigure}{0.4\linewidth}
    \centering

    \includegraphics[height=2.8cm]{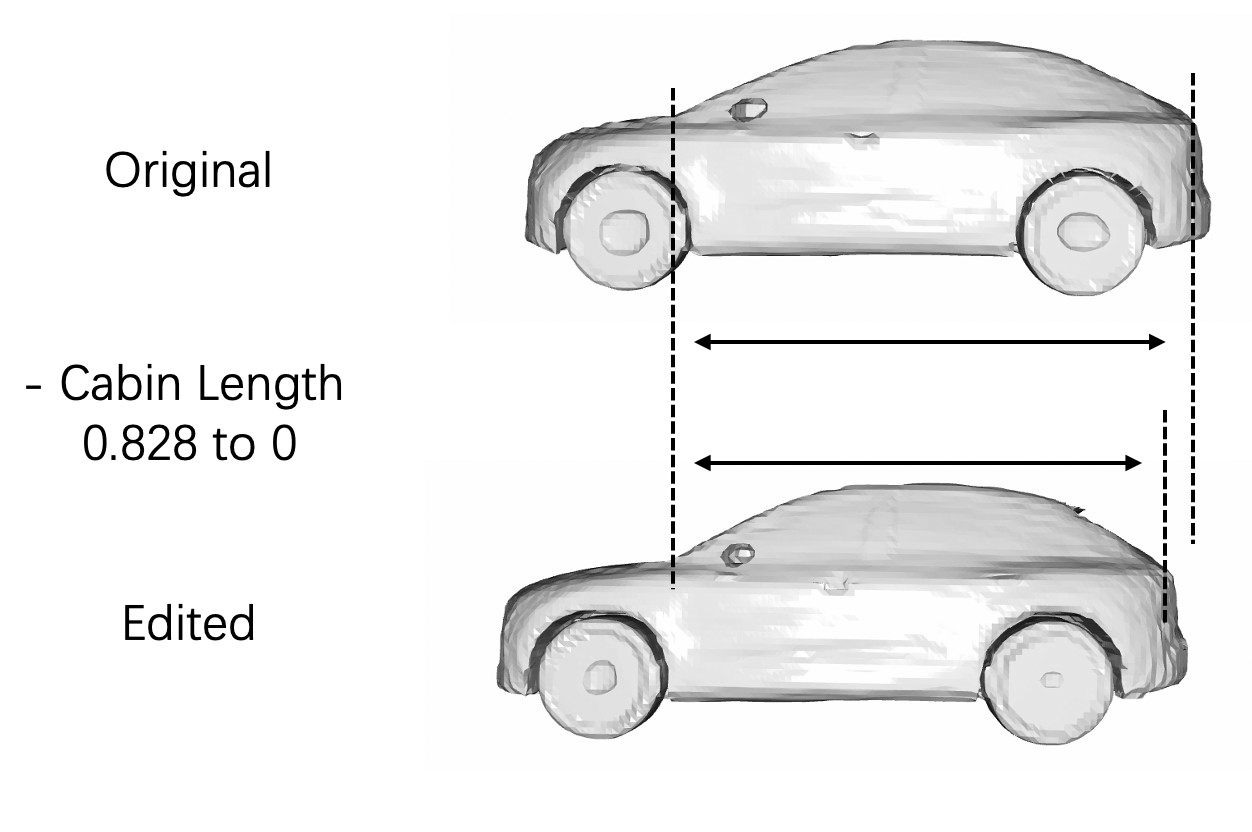}
    \caption{ - Cabin length}
    \label{fig:geo-cabinlength}
  \end{subfigure}

  \vspace{0.3cm} 

  \begin{subfigure}{0.55\linewidth}
    \centering
    
    \includegraphics[height=2cm]{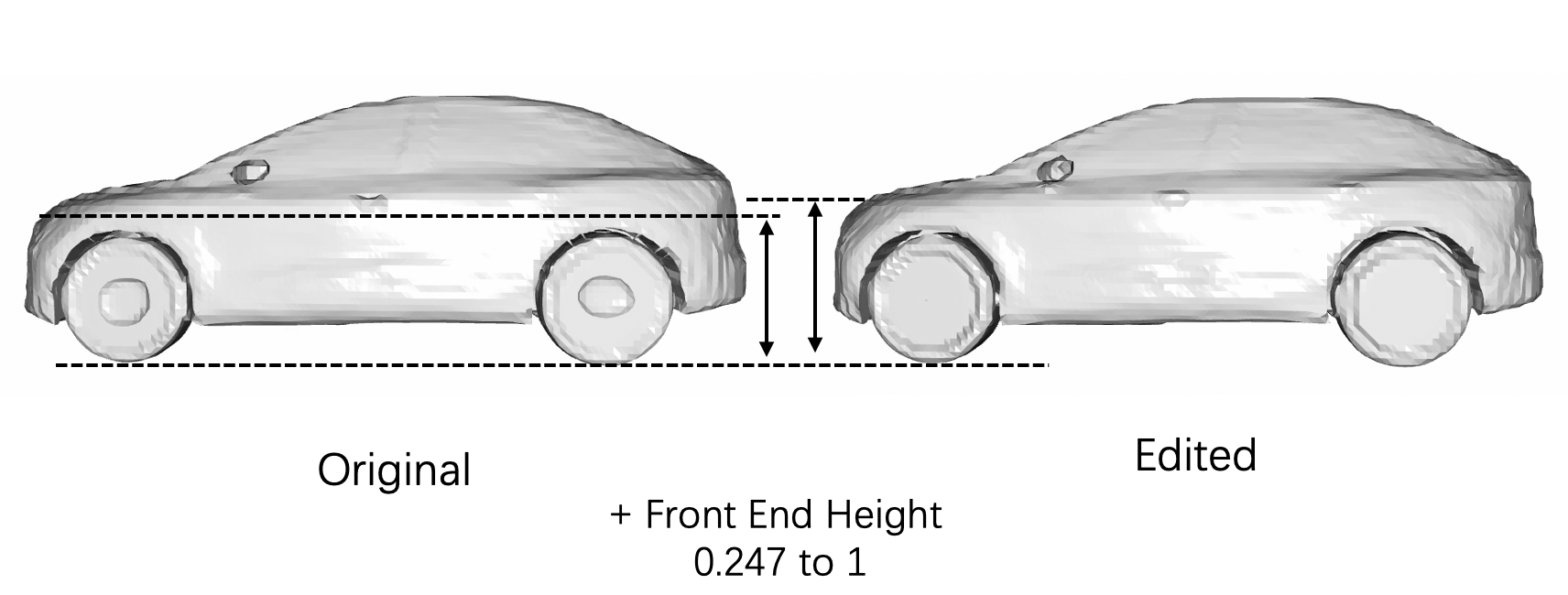}
    \caption{ + Front ending height}
    \label{fig:geo-FR-endheight}
    
  \end{subfigure}
  \hfill
  \begin{subfigure}{0.4\linewidth}
    \centering
    \includegraphics[height=2.5cm]{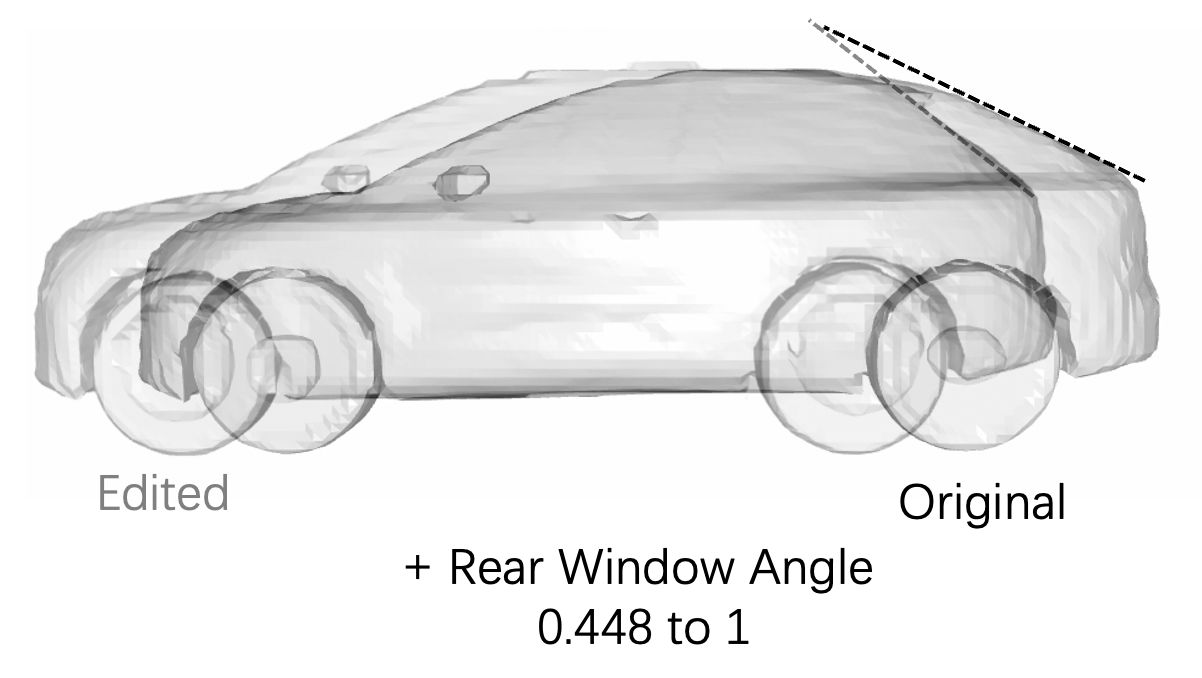}
    \caption{ + Rear window angle}
    \label{fig:RR-winangle}
  \end{subfigure}

  \vspace{0.1cm} 

  \caption{Single Attribute Editing for Geometric Attributes (2)}
  \label{fig:geo-single2}
\end{figure}

\begin{figure}[tb]
  \centering
  \includegraphics[height=4.5cm]{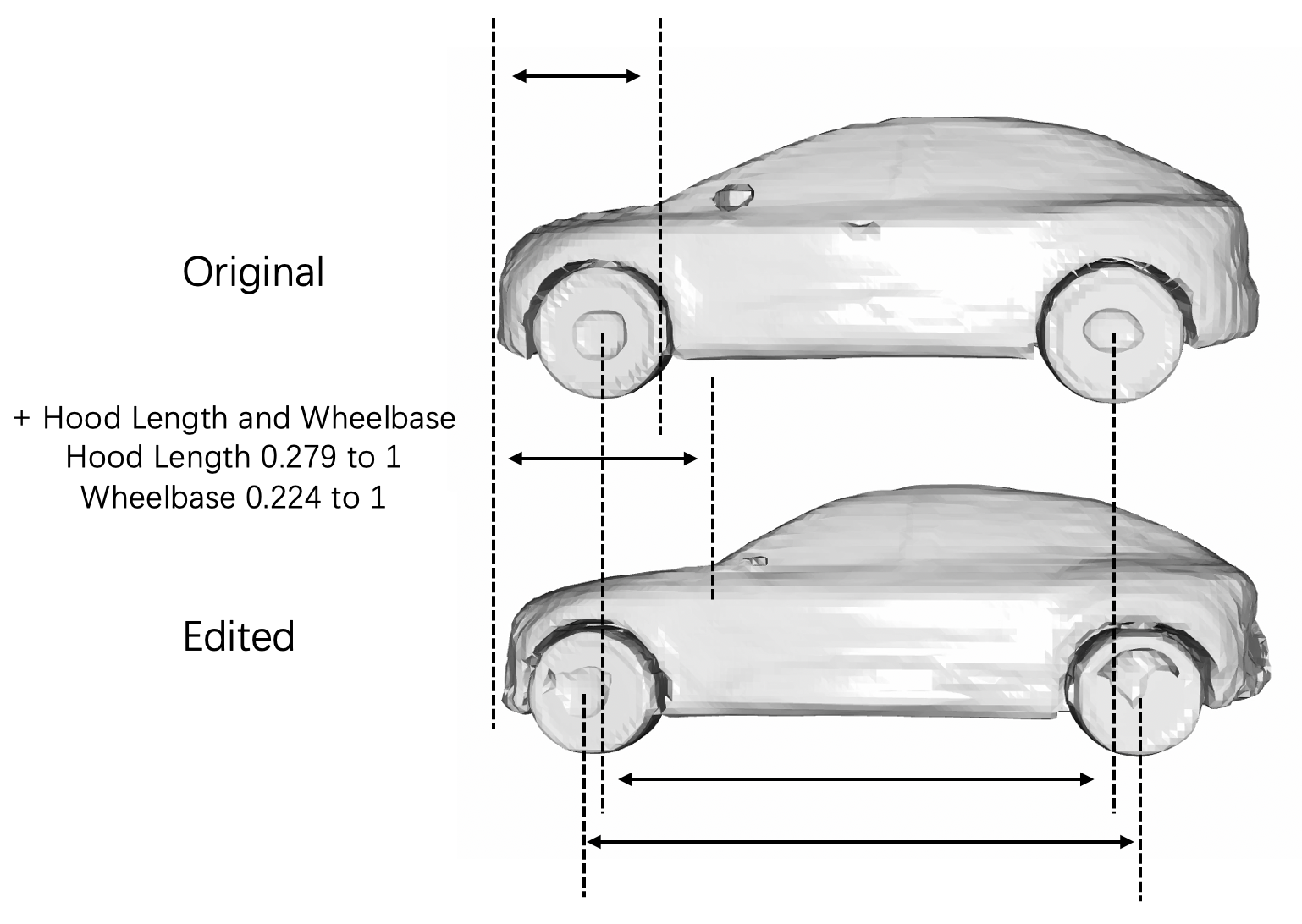}
  \caption{Multi-attributes Editing for Geometry-Related Attributes
  }
  \label{fig:geo-multi}
\end{figure}

\subsubsection{Geometry-Related Attributes Editing Results}
In \cref{fig:geo-single,fig:geo-single2}, we present several geometry-related attribute editing results that are straightforward to demonstrate visually. The numbers displayed below the specified attribute editing types represent the intensity of the attribute modifications (recall that all attribute values have been scaled to take values between 0 and 1). This setting is the same in all of the following figures. Given that both KAN-based and MLP-based editors achieved comparable outcomes in the editing of geometric attributes, we have chosen to display only the results from the KAN-based editor for clarity and brevity.

The results shown in Figs. \cref{fig:geo-single,fig:geo-single2} illustrate the ability of our framework to perform single-variable geometry-related attribute editing, specifically highlighting its capacity to accurately execute the intended attribute transformations without changing the context information of the vehicle. It is evident from the visualizations that our method effectively achieves the desired modifications, denoted by both positive and negative adjustments, while preserving the integrity of the vehicle’s contextual information.

It is important to note that some observers might feel the editing in geometric attributes as limited, as depicted in the figures. We attribute this observation to two fundamental aspects of our methodology. First, a primary objective of our approach is to maintain the original individual characteristics of the vehicle, ensuring that non-targeted attributes remain largely unaffected. Second, our editing constraints are designed to ensure that changes remain within the bounds of the vehicle category, thus preventing the modifications from altering the object’s fundamental category. 

In \cref{fig:geo-multi}, we illustrate the result of simultaneously editing multiple geometry-related attributes, specifically in this case editing both the hood length and wheelbase length. The results from this figure demonstrate our method’s capability to concurrently edit multiple attributes, while maintaining the remaining geometric attributes. The editing result clearly demonstrates that the edited model exhibits an increased hood length and wheelbase, while maintaining the original size of other geometric attributes as much as possible. Furthermore, despite these changes, the overall appearance of the vehicle remains largely consistent. This underscores our framework’s ability to achieve targeted editing without compromising the vehicle’s fundamental design and aesthetic consistency. This suggests that our method could significantly assist automobile manufacturers in making preliminary aesthetic assessments and conducting aerodynamic studies.

\subsubsection{Style-Related Attributes Editing Results}

\begin{figure}[tb]
  \centering
  \includegraphics[height=4.3cm]{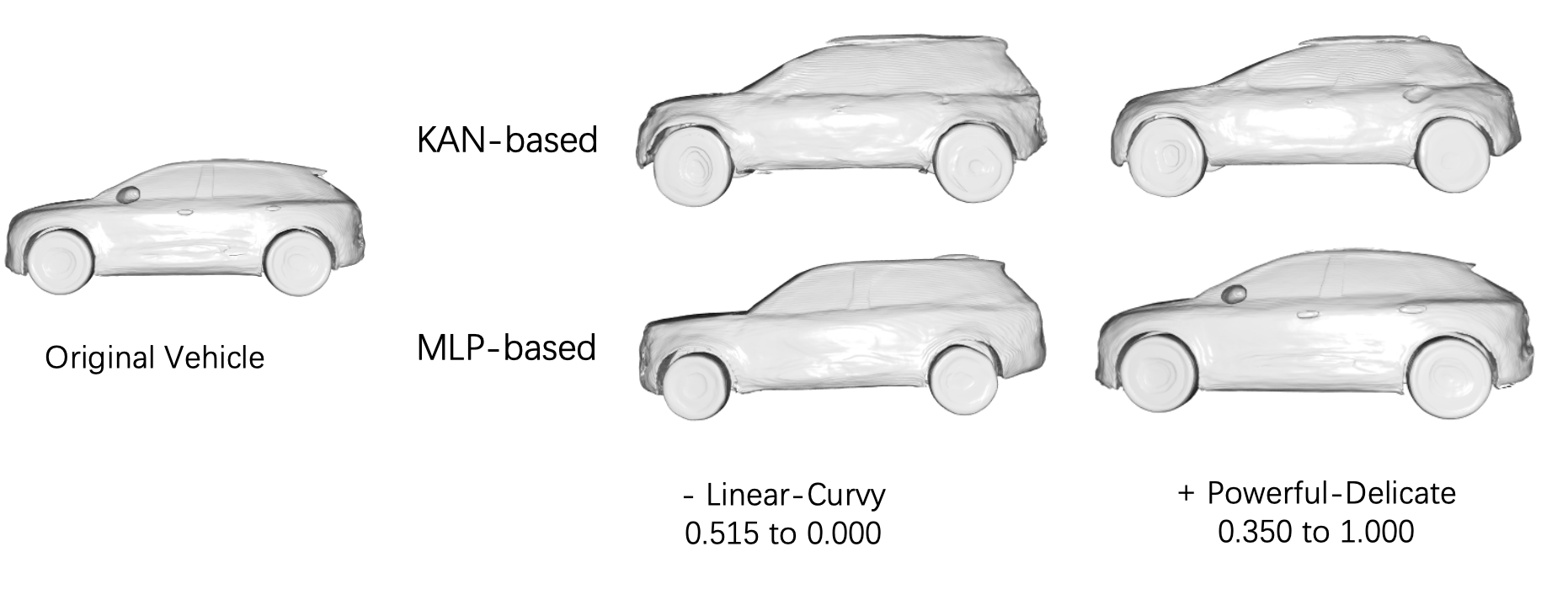}
  \caption{Style attribute editing over several
   predefined pairs of mutually exclusive style attributes (1). Note that, for example, in the pair Linear-Curvy, a value closer to 0 indicates a style more aligned with Linear, whereas a value closer to 1 indicates a style closer to Curvy. Thus, the editing [- Linear-Curvy] is to make the model more closely resemble the Linear style. All style-related labels are normalized to change in the range of [0, 1]. 
  }
  \label{fig:style-editing}
\end{figure}

\begin{figure}[tb]
  \centering
  \includegraphics[height=4.3cm]{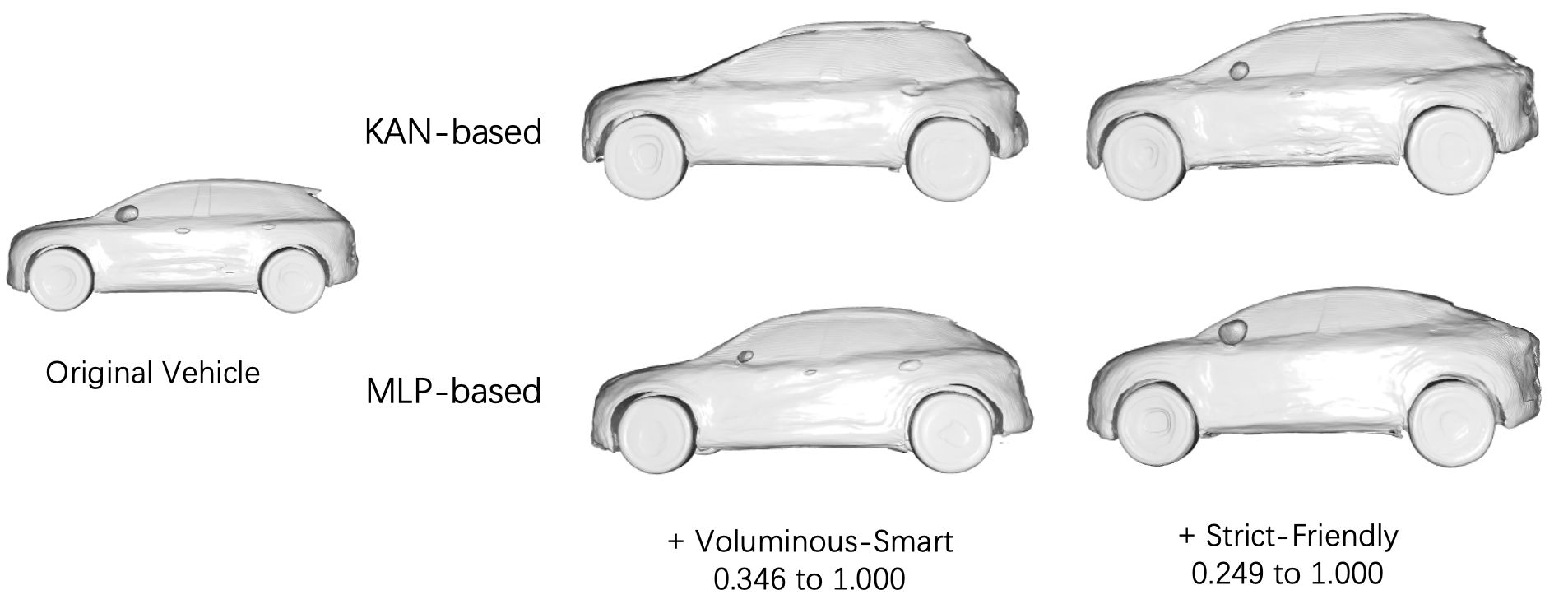}
  \caption{Style attribute editing (2)
  }
  \label{fig:style-editing2}
\end{figure}

In addition to editing the geometric attributes of the vehicles, 
we performed another experiment, in which the regressor was trained on the 14 pairs of predefined stylistic attributes (shown in Table \ref{tab:stylistic attributes}; as in the case of the geometry-related attributes, the style-related labels were normalized to change in the range of [0, 1]). 


This enables our method to edit the style attributes of the vehicles’ appearances. \cref{fig:style-editing} and \cref{fig:style-editing2} illustrates the outcomes of several style-related attribute edits. It should be noted that the editing outcomes for style attributes differ between the KAN-based and MLP-based editors. Both editors fulfill the editing requirements to some extent; this discrepancy likely stems from the editors’ differing interpretations of style attributes. In style-related editing the notation [- Linear-Curvy] means that we are demonstrating the result of enhancing the 'Linear' style attribute (i.e. decreasing the 'Curvy' style attribute), the 'Linear' style attribute is accentuated, which diminishes to some extent the roundness in the design of the original vehicle model, the vehicle starting to look more like a SUV-like vehicle. In editing [+ Power-Delicate], the 'Delicate' style attribute is accentuated, where the vehicle's front and rear windows are angled more sharply, making the cockpit look even smaller, thereby intensifying the perception of Delicate. \cref{fig:style-editing2} shows the outcomes of [+ Voluminous-Smart] and [+ Strict-Friendly].

Given that style attributes are inherently subjective and influenced by the evaluator’s perspective, it is challenging to ensure that such edits align with universal expectations. Moreover, some observers might find the edits between some different style attribute axes looking quite similar. 
The reason for this is that the different style-related axes are not orthogonal to each other, rather exhibiting a certain degree of correlation between them which is inevitable. 

\begin{figure}[tb]
  \centering
  \includegraphics[height=3.5cm]{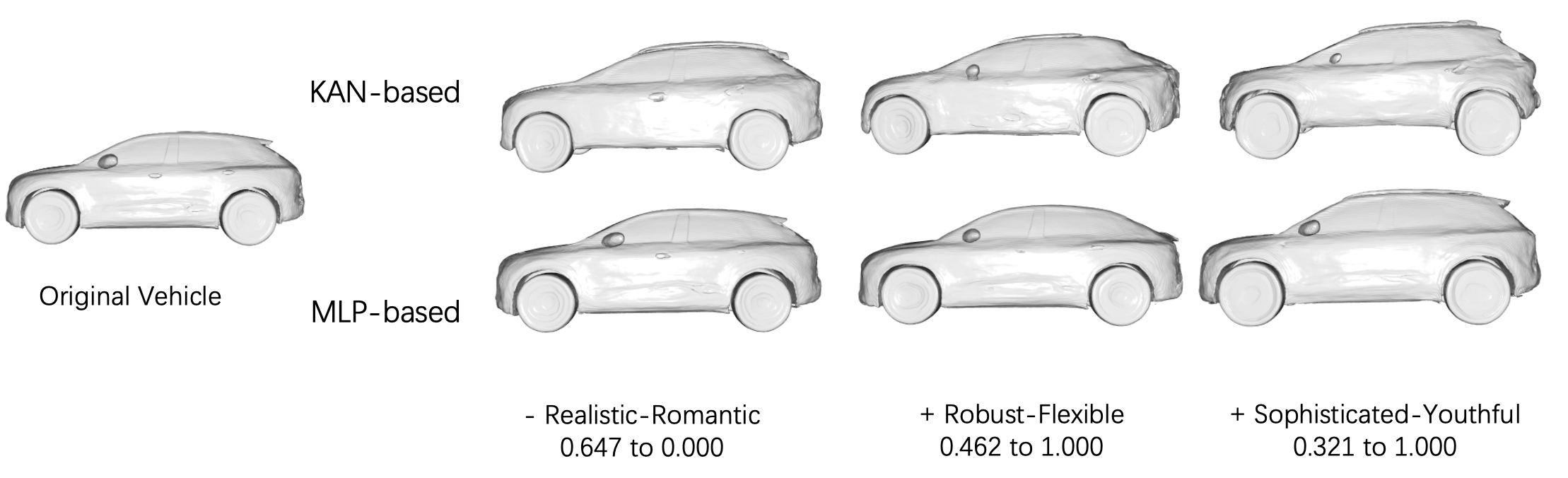}
  \caption{Some other Style attribute editing results
  }
  \label{fig:style-editing3}
\end{figure}

Nevertheless, we believe that overall the style-related attribute editing results also confirm the adaptability of our framework. By gathering data on desired attributes and training a custom-trained regressor, people can effectively tailor edits to 3D models according to their specific needs.

\section{Conclusion}
This paper proposes a framework 
for fine-grained editing of 3D vehicle models in latent space. Utilizing the continuous signed distance function (SDF) with latent code representation of the modified DeepSDF model, our framework not only is able to reconstruct complex 3D vehicle shapes, but also enables continuous fine-grained geometry and style related attribute editing in the latent space. Our approach significantly enhances the precision and flexibility of shape manipulations, enabling users to directly manipulate real 3D objects based on detailed attribute data. Future work will explore extending the current framework to support a broader range of 3D objects and further optimize the exploration efficiency of the latent space.

%
%
\bibliographystyle{splncs04}
\bibliography{main}
\end{CJK}
\end{document}